\documentclass[conference, onecolumn]{IEEEtran}
\IEEEoverridecommandlockouts
\usepackage{cite}
\usepackage{amsmath,amssymb,amsfonts}
\usepackage{graphicx, float}
\usepackage{textcomp}
\usepackage{xcolor}
\usepackage{graphics}
\usepackage[utf8]{inputenc}

\usepackage{csquotes}
\usepackage{subcaption}
\usepackage{multicol}
\usepackage{lipsum}
\usepackage[bookmarks=false]{hyperref}
\usepackage{amsmath,amssymb}
\usepackage{amsfonts}
\usepackage{float}
\usepackage{wrapfig}
\def\BibTeX{{\rm B\kern-.05em{\sc i\kern-.025em b}\kern-.08em
    T\kern-.1667em\lower.7ex\hbox{E}\kern-.125emX}}
    
\usepackage{tabularx}

\usepackage{algorithm}
\usepackage{algpseudocode}
\algblock{Input}{EndInput}
\algnotext{EndInput}
\algblock{Output}{EndOutput}
\algnotext{EndOutput}
\usepackage{authblk}
\usepackage{makecell}

\usepackage{tikz}
\usetikzlibrary{matrix}
\usepackage{xparse}
\usepackage{soul}
\usepackage{float} % Add this package in the preamble
\newcommand*{\affaddr}[1]{#1} % No op here. Customize it for different styles.
\newcommand*{\affmark}[1][*]{\textsuperscript{#1}}
\newcommand*{\email}[1]{\textit{#1}}

\begin{document}

\title{  IoT Malware Network Traffic Detection using Deep Learning and GraphSAGE Models\\}

%%%Deep Learning and GraphSAGE for IoT Malware Network Traffic Detection
% \author[]{}
\author{Nikesh Prajapati\affmark[1],
Bimal Karki\affmark[2], Saroj Gopali\affmark[2], Akbar Siami Namin\affmark[2]\\
\affaddr{Department of Computer Science\affmark[2]},  
\affaddr{Texas Tech University\affmark[2]}
\\
\email{nikeshprajapati7@gmail.com}, \email{\{bimal.karki, saroj.gopali, akbar.namin\}}@ttu.edu
}

\maketitle

\begin{abstract}
%Malware attacks through network traffic are prevalent nowadays primary due to widespread use of IoT devices and their expositions to security threats. Large organizations usually utilize Virtual Private Network (VPNs) to use or transfer their sensitive data. Basic security mechanisms, such as encryption and multi-factor authentication, are used to protect data. However, preventing network intrusion launched by attackers yet remains a challenging problem. 
This paper intends to detect IoT malicious attacks through deep learning models and demonstrates a comprehensive evaluation of the deep learning and graph-based models regarding malicious network traffic detection. The models particularly are based on GraphSAGE, Bidirectional encoder representations from transformers (BERT), Temporal Convolutional Network (TCN) as well as Multi-Head Attention, together with Bidirectional Long Short-Term Memory (BI-LSTM) Multi-Head Attention and BI-LSTM and LSTM  models. The chosen models demonstrated great performance to model temporal patterns and detect feature significance. The observed performance are mainly due to the fact that IoT system traffic patterns are both sequential and diverse, leaving a rich set of temporal patterns for the models to learn. Experimental results showed that BERT maintained the best performance. It achieved 99.94\% accuracy rate alongside high precision and recall, F1-score and AUC-ROC score of 99.99\% which demonstrates its capabilities through temporal dependency capture. The Multi-Head Attention offered promising results by providing good detection capabilities with interpretable results. On the other side, the Multi-Head Attention model required significant processing time like BI-LSTM variants. The GraphSAGE model achieved good accuracy while requiring the shortest training time but yielded the lowest accuracy, precision, and F1 score compared to the other models.

% The GraphSAGE model achieves the lowest accuracy The training time evaluation established an accuracy-computational resources trade-off that positions BERT as the most suitable option for real-time deployment in IoT systems. 
\end{abstract}


\section{Introduction}
% The widespread use of IoT devices and services has made network traffic malware attacks more common because IoT devices remain unusually exposed to security threats. 

Network traffic has experienced significant growth due to the rise of Internet of Things (IoT) devices and services. While the use of IoT-based devices is on rise, the security of these devices are also of major concerns. As a motivating example, IoT faces malware attacks due to its dynamic and distributed nature, which makes it highly susceptible to cyberattacks. Most IoT devices operate without sufficient security protections so they present easy targets for attackers to develop malware, launch botnets attacks through Distributed Denial-of-Service \(DDoS\) campaigns \cite{abosata2021internet}.

Although IoT devices are designed and deployed for various purposes, most of these devices share one common characteristic: they are connected to networks, making them prime targets for network-based attacks. Hence, analyzing the network activity of these devices is crucial to determine if they have been compromised, as network activity is a critical metric in classifying processes as malicious. Many malicious applications attempt to connect to remote servers or use them for attacks, with demonstrated behaviors such as Distributed Denial of Service (DDoS) attacks, transferring stolen sensitive data, or establishing remote connections to upload or download files. These malicious actions may generate significant network traffic that can be analyzed to identify malicious applications. For example, in a DDoS attack, a large number of requests is sent to a server, overwhelming it and thus causing a crash. The patterns created by such activities can be detected using machine learning techniques, enabling dynamic malware detection by leveraging behavioral analytics in real-time.

This paper compares the performance of six deep learning models namely Long Short-Term Memory (LSTM), Bidirectional Long Short-Term Memory (BI-LSTM), Temporal Convolutional Network (TCN), Multi-Head Attention, BI-LSTM with Multi-Head Attention and Bidirectional Encoder Representations from Transformers (BERT) with the goal of detecting malicious network traffic. In addition to deep learning models, the paper also evaluates the performance of a graph-based model, called GraphSAGE \cite{hamilton2017inductive}. The research uses the Malware Network traffic Dataset \cite{sebastian_garcia_2020_4743746} that includes labeled data for both malicious and benign processes. The dataset contains \texttt{1,008,748} data points with \texttt{23} columns (i.e., features). The key contributions of this research are as follows:

\begin{enumerate}
    \item This research compares six deep learning models including LSTM, BI-LSTM, Multi-Head Attention, BI-LSTM with Multi-Head Attention, TCN, and BERT for modeling malicious network data and evaluates their performance and effectiveness in malware.
    \item The result demonstrates that while GraphSAGE requires minimal training time, it underperformed classification and detection compared to the other models studied.% in this work.
    
    \item The study explores the impact of the BERT Model for malware detection by analyzing its effectiveness in capturing temporal dependencies with the highest AU-ROC of 99.99\%.  
    
    % \item The research utilizes network features such as originating IP address, destination IP address, payloads, and duration to train the model. These characteristics record both behavioral, as well as structural aspects of network traffic so that deep learning models can identify benign and malicious activities efficiently. 
    
    \item  TCN Model exhibits outperformed the other models by achieving 99.30\% accuracy and 99.70\% AUC-ROC score for detection of malicious network traffic.

\end{enumerate}

The rest of the paper is organized into the following sections. The related work is discussed in Section \ref{sec:related}. %The technical overview of deep learning models is presented in Section \ref{sec:overview}. 
The methodology Section \ref{sec:methodology} presents the information of the algorithms employed in the experiment. The experimental Section \ref{sec:expriment} discusses the information on dataset and deep learning architectures. Section \ref{sec:modelarchitecture} presents the model architectures. Section \ref{sec:result} presents the results from the experiment. Section \ref{sec:conclusion} concludes the paper along with future research directions.

\section{Related work}
\label{sec:related}

% \begin{figure*}[!htb]
%     \centering
%     \includegraphics[width=\textwidth,  height = 4cm]{figure/dataflow.png}
%     \caption{Data flow combining data preprocessing, models and performance metrics.}
%     \label{fig:methodology}
% \end{figure*}
% This section discusses existing research on the use of machine-learning algorithms for malware detection in IoT devices. 

Arora et al. \cite{arora2014malware} propose a novel approach to detect android-based, remotely controlled malware. The approach is based on the network traffic features by building a rule-based classifier instead of relying on static and dynamic analysis, which either cannot be trusted or is too resource intensive. %The authors collected and analyzed 27 malware families to run on an Android emulator, and used 16 network traffic features such as Average Packet Size (APS), average number of packets sent and received per flow. A threshold was applied to classify the risk level (i.e., low, medium, and high) by analyzing and comparing the extracted features with existing malware traffic ranges. 
They trained and tested rule-based classifier in 43 malware traffic and 5 normal mobile traffic samples. The authors achieved an accuracy of 93.75\% for the malware traffic  classification models.

Shabtai et al. \cite{shabtai2014mobile} propose a detection system that monitors malicious behavior in a network through mobile applications.  Initially, network traffic-related data, such as source and destination IP addresses, protocols, and size of packets are generated by mobile applications. Next, relevant features such as communication between various IP addresses, average connection duration, and packet size are extracted from the data that would portray the normal network behavior of mobile applications. Then, local and collaborative models are created, where the local model gets a traffic pattern from the user applications, and the collaborative model analyzes the network traffic from multiple users to derive a normal network behavior for the application. %If there is a deviation after the comparison of the model-induced behavior against the network traffic of the mobile application, the detection system flags the network as malicious. Deviations of up to 20–25 percent were observed in tests with the same application versions. Authors report deviations of 60\%  or more in the applications containing injected malware.

Lui et al. \cite{liu2019integrated} explore leveraging both static and dynamic analyses of malware. The authors integrated the Interactive Disassembler (IDA) and the Cuckoo sandbox to extract the static and dynamic behavior of malware. They furthered the static analysis by extracting the functional graph, the behavior of the host, and the behavior of the network. The dynamic analysis was then performed using network traffic, where port number statistics were analyzed. The authors were able to successfully extract and detect Mirai malware behavior and several variants by integrating the IDA disassembler and Cuckoo sandbox. %Mirai operates as a highly destructive IoT malware that infects linked devices before converting them into bot-operated systems that enable remote control access. The authors were also able to analyze the evolution and spread of malware in the IoT through large-scale network traffic analysis. 

Jeon et al. \cite{jeon2020dynamic} explored Convolutional Neural Network (CNN) models for dynamic analysis of IoT malware. %Malware samples were executed in a controlled cloud environment, and behavior samples were analyzed. 
The authors used IDA Pro analysis tools to generate assembly code from the binary file and debugged it remotely on the target file to identify how the code works. Various information such as memory, network, system call, Virtual File System (VFS), and process features were extracted from the file. The extracted behavioral data was used to create the visual representations, such as images. %Finally, the CNN model was used to process images containing behavioral data to distinguish between malicious and benign networks. 
The authors used 561 files as testing dataset to achieve a detection accuracy of 99.28\% for the dynamic analysis of IoT malware detection.

Gopali et. al \cite{gopali2022deep} performed a deep learning-based time series analysis for anomaly detection in IoT. The authors utilized the SWaT dataset and compared their results using variations of LSTM-based models and CNN-based TCN models. The research indicates that CUDA Deep Neural Network (CuDNN-LSTM)  models exhibit exceptional accuracy when evaluating in various timestamps with the lowest RMSE values. 
% at 30, 20, 15, 10, 7, and 5 minutes as their RMSE values reach 0.035, 0.041, 0.043, 0.041, 0.043, 0.049. 
% The RMSE values produced by TCN-based models measure 0.065, 0.066, 0.061, 0.057, 0.058, and 0.069 while 
The TCN-based models also achieved the lowest RMSE values while their training times remain significantly shorter than those of CuDNN-LSTM models. The author observed that TCN performs relatively well in detecting anomalies with relatively low root mean square error (RMSE) with short training time. 

\section{Model-Building Algorithms}
\label{sec:methodology}

The section presents the algorithm for building the models that studied during the experiment in the study. 
% The process starts with data preprocessing before performing dataset split, followed by the model architecture definition and activation selection.  The prepared model receives a test dataset for binary classification evaluation to confirm its operational efficiency.

% The model learning occurs through sequential forward propagation, which then calculates binary cross-entropy loss before applying parameters modifications using Adam backpropagation.

\begin{algorithm}
\caption{Training Deep Learning Models for Binary Classification}
\label{algo:deeplearing}
\begin{algorithmic}[1]
\State \textbf{Input:} Dataset $D$ , model $M$,  epochs $E$, batch size $B_s$ %= \{(x_i, y_i)\}_{i=1}^{N}$
\State \textbf{Output:} Trained model $M$

\State Preprocess, features extraction and normalize dataset $D$
\State Split dataset into training, validation, and test sets
\State $M$ = LSTM $\vert$ BI-LSTM $\vert$ TCN $\vert$ Multi-Head Attention $\vert$ BI-LSTM Multi-Head Attention 
\State Model $build$ ($M$)
% \State  $binary cross-entropy$ as loss function
% \State  $Adam$ optimizer 

\For{epoch $= 1$ to $E$}
    \For{each batch $\{(x_b, y_b)\}$ of $B_s$ from training set}
        % \State Perform forward pass
        \State Compute loss $L$ using binary cross-entropy:
        \State Perform backpropagation: Compute gradients %$\nabla L$
        \State Update model parameters using Adam
        % \[ \theta \leftarrow \theta - \eta \nabla L \]
    \EndFor
    \State Evaluate model on validation set
\EndFor

\State Return trained model $M$
\end{algorithmic}
\end{algorithm}

Algorithm \ref{algo:deeplearing} lists a structured approach to develop deep learning models that include LSTM together with BI-LSTM and TCN and Multi-Head Attention for binary classification purposes.

\begin{algorithm}
\caption{Fine-tuning Pretrained BERT for Binary Classification}
\label{alg:bert}

\begin{algorithmic}[1]

\State \textbf{Input:} Pretrained BERT model $B$, labeled dataset $D$, tokenizer $T$, learning rate $\eta$, epochs $E$, batch size $B_s$
\State \textbf{Output:} Fine-tuned BERT model $B$

\State Initialize BERT model $B$ with pretrained weights
\State Add a classification head with softmax activation
\State Tokenize input texts using tokenizer $T$
\State  \texttt{MalwareDataset} to handle encoding and labels

\State Initialize model with training arguments
\For{epoch $= 1$ to $E$}
    % \For{each batch b such as b =$\{(x_b, y_b)\}$ of size $B_s$ from $D$}
    \For{each batch $b = \{(x_b, y_b)\}$ of size $B_s$, where $x_b$ and $y_b$ are the input sequences and corresponding labels, respectively}

        \State Tokenize batch input $x_b$ and obtain encoded inputs.
        % \State Pass embeddings through BERT 
        \State Compute logits using classification head
        \State Compute loss using binary cross-entropy:
        % \[ L = -\frac{1}{B_s} \sum_{i=1}^{B_s} \left[ y_i \log \hat{y}_i + (1 - y_i) \log (1 - \hat{y}_i) \right] \]
        \State Compute gradients $\nabla L$ w.r.t model parameters
        \State Update model parameters using Adam optimizer:
        % \[ \theta \leftarrow \theta - \eta \nabla L \]
    \EndFor
\EndFor

% \State Tokenize test inputs using $T$
% \State Create test dataset \texttt{MalwareDataset}
% \State Perform inference using the trained model
% \State Obtain predictions using \texttt{argmax} on model output logits

\State Return fine-tuned model $B$
\end{algorithmic}
\end{algorithm}

Algorithm \ref{alg:bert} presents the steps for the BERT model. The model performs binary classification through pre-trained BERT model fine-tuning after labels are encoded and the text is tokenized during preprocessing which prepares data for conversion into PyTorch tensors. 

% The model training process occurs through a Trainer setup that utilizes specified arguments for learning rate scheduling and weight decay to perform backpropagation based updates on its parameters.
% As a last step the model conducts inference through a test data pipeline that starts with tokenization followed by prediction generation through the classification head.

% It connects these nodes with edge that includes detailed connection statistics and computes aggregated measurements such as failure rate and average packet.

\begin{algorithm}
\caption{Building and Training The GraphSAGE Model}
\label{alg:graphsage}
\begin{algorithmic}[1]
\State \textbf{Input:} Dataset \textit{D} with network traffic records
\State \textbf{Output:} Trained GraphSAGE model

\State Initialize graph $G$, mapping $Map$, and index counter $i$
\For{each record $r$ in \textit{D}}
    \State \textbf{Update Nodes:} 
    \State \quad Add source and destination IPs to $G$ and $Map$
    % \State \quad Accumulate node metrics (duration, bytes, packets, ..)
    \State \quad Add edge $(src,dst)$ with (duration, ports, bytes, ...)
\EndFor
\For{each node in $G$}
    \State Compute aggregated metrics (e.g., unique connections)
    \State Apply log-transform to select attributes
\EndFor
\For{each edge in $G$}
    \State Compute derived features (e.g., average packet size)
\EndFor
\State Convert $G$ into a PyG data with node, edge, and labels
\State Split the dataset into training and testing sets
\State Initialize the GraphSAGE model and training parameters
\For{epoch = 1 to $N$}
    \State \textbf{Training Loop:} 
    \State \quad For each batch from a neighbor sampler, perform:
    \State \quad \quad Computer loss, backpropagation, weight update
    \State \quad \quad Record performance metrics (loss, accuracy)
\EndFor
\State \Return Trained GraphSAGE model
\end{algorithmic}
\end{algorithm}

Algorithm \ref{alg:graphsage} presents the steps of GraphSAGE model data processing and training. Initially, the algorithm builds the traffic logs into a network graph by turning unique IP-address into nodes and aggregating communication statistics. The $Map$ creates a mapping from IP addresses to unique integer indices.  The node attributes are then transformed (using log operations) to a PyG data structure for further processing. The algorithm then builds a GraphSAGE model on this data using neighbor sampling to train iteratively a classification.

\section{Experimental Procedure}
\label{sec:expriment}

% The section presents the description of a dataset, the data pre-processing steps, the deep learning models and their architecture, and the assessment metrics which were employed in the experiment. 

% Figure \ref{fig:methodology} demonstrates the end-to-end pipeline that was implemented in this research. 

\subsection{Dataset and Pre-processing }
The dataset studied in this research is the labeled Aposemat IoT-23, which has been captured by the Stratosphere Laboratory, AIC Group, FEL, and CTU University in the Czech Republic \cite{sebastian_garcia_2020_4743746}. First published in January 2020, the dataset includes network traffic captures from 2018 to 2019. Its primary aim is to provide a large-scale resource to aid researchers in developing machine learning algorithms for IoT security. The dataset contains $1,008,748$ data points with 23 features. The dataset is well-balanced, with 53.5\% malicious and 46.5\% benign instances. It consists of a total of 23 features out of which we selected only 17 features as follows: {\tt ts}, {\tt id.orig\_h}, {\tt id.orig\_p}, {\tt id.resp\_h}, {\tt id.resp\_p}, {\tt proto}, {\tt duration}, {\tt orig\_bytes}, {\tt resp\_bytes}, {\tt conn\_state}, {\tt missed\_bytes}, {\tt history}, {\tt orig\_pkts}, {\tt orig\_ip\_bytes}, {\tt resp\_pkts}, {\tt resp\_ip\_bytes}, and {\tt label}. These selected features capture both behavioral and structural characteristics  of network traffic so that deep learning models can differentiate between benign and malicious activities efficiently. The remaining six features, i.e., {\tt uid}, {\tt service}, {\tt local\_orig}, {\tt local\_resp}, {\tt tunnel\_parents}, and {\tt detailed-label} were dropped during the data pre-processing phase.

% The features \texttt{uid, service, local\_orig, local\_resp and tunnel\_parents} were dropped because they served as special identifiers and incomplete or insufficient behavior patterns to discriminate between benign and malicious activities. Additionally, the \texttt{detailed\-label} was removed because it protected against data leakage and delivered broader applicability.

In the data pre-processing step, all missing values such as '-' were replaced as \texttt{NA}. The IP addresses were converted to integers, and both numerical and categorical data were identified. Categorical values were One-Hot Encoded, and the data normalization was performed to enhance model performance. Finally, the data were split into 70\% and 30\% as training and testing sets, respectively. The model utilized 20\% of training data to validate the model during training. 

\iffalse
Figure \ref{fig:data_dis} depicts a network subgraph sampling process which extracts related node and edge connections from a bigger diameter graph structure that features entities and their related interactions as nodes connected through edges. 

\begin{figure} [ht]
    \centering
    \includegraphics[width=\linewidth]{figure/graph_vis.png}
    \caption{Connected Network Graph of Source IP to destination IP Address}
    \label{fig:data_dis}
    \vspace*{-0.20in}
\end{figure}
\fi

% The visualization includes edge labels that show the communication protocol and byte transfer numbers to give additional information regarding network connections.

During the training stage of building the GraphSAGE model, the dataset was reduced to $602,833$. The reduction mainly occurs through the code process of unifying multiple duplicated records together. The initial data contained $1,008,748$ records of pairs but many of these entries linked the same IP addresses to each other. The architecture of NetworkX \cite{hagberg2008exploring} adds each IP address as a node to Graph only once and every edge between two nodes appears once at most.  A particular IP cannot exceed one representation in the visualization although it might show up across multiple data rows. The {\tt G.add\_edge(src, dst, ...)} function updates the existing edge between two nodes rather than creating a duplicate edge when they share identical nodes. The system combines multiple identical rows between two hosts into a single graphical edge. Graph aggregation reduces the duplicate communications endpoints because several rows reference the same endpoints, thus creating 602,833 unique connections, which can be seen in the confusion matrix of the GraphSAGE Model as shows in Figure \ref{fig:graphsage_conf}.
% The system creates a new node from any IP address that appears once from either the source or destination fields.
\subsection{Performance Metrics}
\subsubsection{Confusion Matrix}
During the experiment, the confusion matrix was calculated including  precision, recall, F1-score, and accuracy metrics. The number of true positive (TP), false positive (FP), true negative (TN), and false negative (FN) predictions were determined from the confusion matrix in binary classification which then was used to compute the precision, recall, and F1$\-$ score.

% \begin{equation}
% Precision = \frac{TP}{TP + FP}
% \end{equation}

% \begin{equation}
% Recall = \frac{TP}{TP + FN}
% \end{equation}

% \begin{equation}
% F1\text{-}Score = 2 \cdot \frac{Precision \cdot Recall}{Precision + Recall}
% \end{equation}

\subsubsection{ROC}
The Receiver operating characteristics (ROC) \cite{fawcett2006introduction} is the most popular evaluation metric for classification problems where the True Positive Rate (TPR) is plotted against the False Positive Rate. The AUC value ranges from 0 to 1, where values of 1 indicate perfect classification performance, and values of 0.5 indicate a random classifier.

% \begin{equation}
% TPR = \frac{TP}{TP + FN}
% \end{equation}

% \begin{equation}
% FPR = \frac{FP}{FP + TN}
% \end{equation}

% The Area Under Curve (AUC) is a single value used to determine the performance of the model. 

\section{Deep learning Model Architectures}
\label{sec:modelarchitecture}

The deep learning models utilized TensorFlow %\footnote{https://www.tensorflow.org/ } 
as the backend. BERT model utilized the pre-trained BertForSequenceClassification from Hugging Face Transformers library %\footnote{https://huggingface.co/} 
which operated through PyTorch \cite{paszke2019pytorch}. The LSTM  BI-LSTM, TCN and Multi-Head Attention and BI-LSTM  with Multi-Head Attention models used Keras \cite{chollet2015keras} with TensorFlow as their backend engine. The models, except BERT, utilized Adam \cite{kingma2014adam} as an optimizer and \texttt{binary\_crossentropy} as a loss function. The PyTorch Geometric library \footnote{https://pyg.org/} employed for the GraphSAGE model.

Table \ref{tab:dl_arch} reports the number of layers, units, and dropout values that have been set during the model training in the experiment. 
All models were trained on 20 epochs with 125 batch sizes. The output layer contains the sigmoid activation function across all models. The training process for BERT utilized Transformer's \texttt{Trainer} API during which the following training arguments were employed: epoch= 0.05 batch size 4, weight decay 0.01, logging every 10 steps, and a warm-up phase of 100 steps.

\begin{table}[t]
    \centering
    \begin{tabularx}{0.65\linewidth}{|l|X|}
        \hline
        \textbf{Model} & \textbf{Layers} \\
        \hline
        BI-LSTM & \makecell[l]{- BI-LSTM(unit = 35)} \\
        
        Multi Head Attention & \makecell[l]{- Multi-HeadAttention(num\_heads = 4, key\_dim = 16)\\
                                             - Dropout (0.01)\\
                                             - GlobalAveragePooling1D()} \\
        \hline
        TCN & \makecell[l]{- TCN (unit = 32, dilations = [8, 64])\\
                           - Dropout (0.01)} \\
        \hline
        LSTM & \makecell[l]{- LSTM (unit = 32)\\
                            - Dropout (0.01)} \\
        \hline
        BI-LSTM (with Embedding) & \makecell[l]{- Embedding\\
                                                 - BI-LSTM(unit = 35)\\
                                                 - Dropout (0.01)} \\
        \hline
        BERT & \makecell[l]{- BertForSequenceClassification\\
                            - epochs = 0.05, batch\_size = 4\\
                            - weight\_decay = 0.01} \\
        \hline
        GraphSAGE & \makecell[l]{- GraphSAGE(in\_channels, 256, out\_channels)\\
                                 - weight\_decay = 1e-6} \\
        \hline
    \end{tabularx}
    \caption{Summary of Model Architectures}
    \label{tab:dl_arch}
    \vspace*{-0.15in}
\end{table}

In the GraphSAGE modeling, initially NetworkX was used for graph construction, creation and identification relationships between nodes and edges. Nodes are represented as IP addresses for the host and destination, while the edge weight is the duration of the network session from the host IP to destination IP. Nodes and Edge attributes are crucial because they add detailed information to the relationship in a graph. The node attributes that are considered for the graph in a network session include total bytes transferred, total packets transferred, number of unique connections made, failed connection rate, average size of packets, and the number of unique protocols. Edge attributes are the duration of the network session, duration per packet, number of rejected connections, source and destination ports, total number of packets and bytes transferred. All the numerical data are log transformed, while categorical data such as protocols, connection states, labels, and history are label encoded. 

The NetworkX Graph is then converted to a PyTorch Geometric (PyG) data object, using all the node and edge attributes mentioned above that are passed as the input features to the GraphSAGE to train GNN. GraphSAGE uses SAGEConv library which is a PyTorch geometric layer for GraphSAGE-based message passing. To maintain consistency, the model was trained for 20 epochs and 125 batch sizes.

\section{Results}
\label{sec:result}

\subsection{Model Performance}

Table \ref{table:class_report} reports the results achieved by each model that were measured using Precision, Recall, F1-score, Accuracy, and AUC-ROC metrics. A glance at the table indicates that all models perform comparatively. BERT model achieve the highest precession, recall, F1 score and accuracy of 99.94\%. The model also achieved the highest ROC value of 99.99\% among the models. All the model achieve the accuracy over 99\% but the BERT model outperformed sightly the others.

% The experiment evaluates Six deep learning models i.e. BERT, TCN, Multi-Head Attention, BI-LSTM Multi-Head Attention, BI-LSTM, and LSTM for the detection of malicious network activity.

\begin{table*}[h]

    \centering
     \scalebox{1.0}{
    \begin{tabular}{|l|c|c|c|c|c|}
       
        \hline
         \textbf{Model} 
         % & \textbf{Average} & \textbf{Average} &  \textbf{Average} &  \textbf{Average} & \textbf{Average}\\ 
        
        & \textbf{Precision (\%)} & \textbf{Recall (\%)} & \textbf{F1-score (\%)} & \textbf{Accuracy (\%)} &\textbf{ROC (\%)}\\ \hline
       BERT  & \bf 99.94 & \bf 99.94 & \bf 99.94 & \bf 99.94 & \bf 99.99 \\ \hline
        TCN  &  99.31 &  99.30 &  99.30 &  99.30 &  99.70 \\ \hline
        Multi Head Attention  & 99.29 & 99.28 & 99.28 & 99.28 &99.50 \\ \hline
        BI-LSTM Multi-Head Attention & 99.30 & 99.29 & 99.29 & 99.29 &99.40 \\ \hline        
        BI-LSTM & 99.02 & 99.01 & 99.01 & 99.01 &99.50 \\ \hline
        LSTM & 99.01 & 98.99 & 98.99 & 98.99 &99.50 \\ \hline
        GraphSAGE & 97.21 & 97.16 & 97.19 & 97.16 &97.70 \\ \hline
       
    \end{tabular}
    }
    \caption{Classification report of Models from Experiment}
    \label{table:class_report}
    \vspace*{-0.20in}
\end{table*}

TCN stands second by reaching 99.30\% accuracy as well as a precision of 99.31\%, recall of 99.30\%, F1-score of 99.30\%, and an AUC-ROC score of 99.70\%. The TCN model has demonstrated that it can separate network traffic between malicious and benign sessions exceptionally well. Multi-Head Attention yields an accuracy of 99.30\% which is equivalent to the TCN, but demonstrates a slightly reduced AUC-ROC value of 99.50\% which signifies its capacity to detect dependencies within network traffic data.  

An implementation of BI-LSTM with a Multi-Head Attention configuration achieved comparable results, reaching 99.30\% accuracy and 99.40\% AUC-ROC performance. When the BI-LSTM model is used alone, it reaches an accuracy level of 99.01\% and an AUC-ROC value of 99.50\% whilst showing a slow decrease in its performance compared to its attention-augmented version. The study suggests that the LSTM model exhibited the second worst performance among the considered methods, with an accuracy of 98.99\% and an AUC-ROC value of 99.50\%. However, it still demonstrated a strong ability to identify temporal patterns. The GraphSAGE model performance achieved the lowest in precision of 97.31\%,  recall of 97.16\% , F1-Score of 97.19\%  and accuracy of 97.16\%. The model also achieved the lowest AUC-ROC value of 97.75\% compared to other models.  

The experimental results indicate that BERT is highly effective  for malware detection. Moreover, TCN model excels at recognizing temporal dependencies within network traffic data.  The lowest performance of GraphSAGE derives from its weak capability to identify crucial temporal relationships and data sequence patterns, which network traffic data analysis requires for malware detection purposes.

% The multi-head attention-based models display competitive performance based on the results, which further demonstrates the value of attention mechanisms for enhancing classification outcomes.
\subsection{Model Training Time}

\begin{figure}[t]
    \centering
    \includegraphics[width=0.8\textwidth, height=7cm]{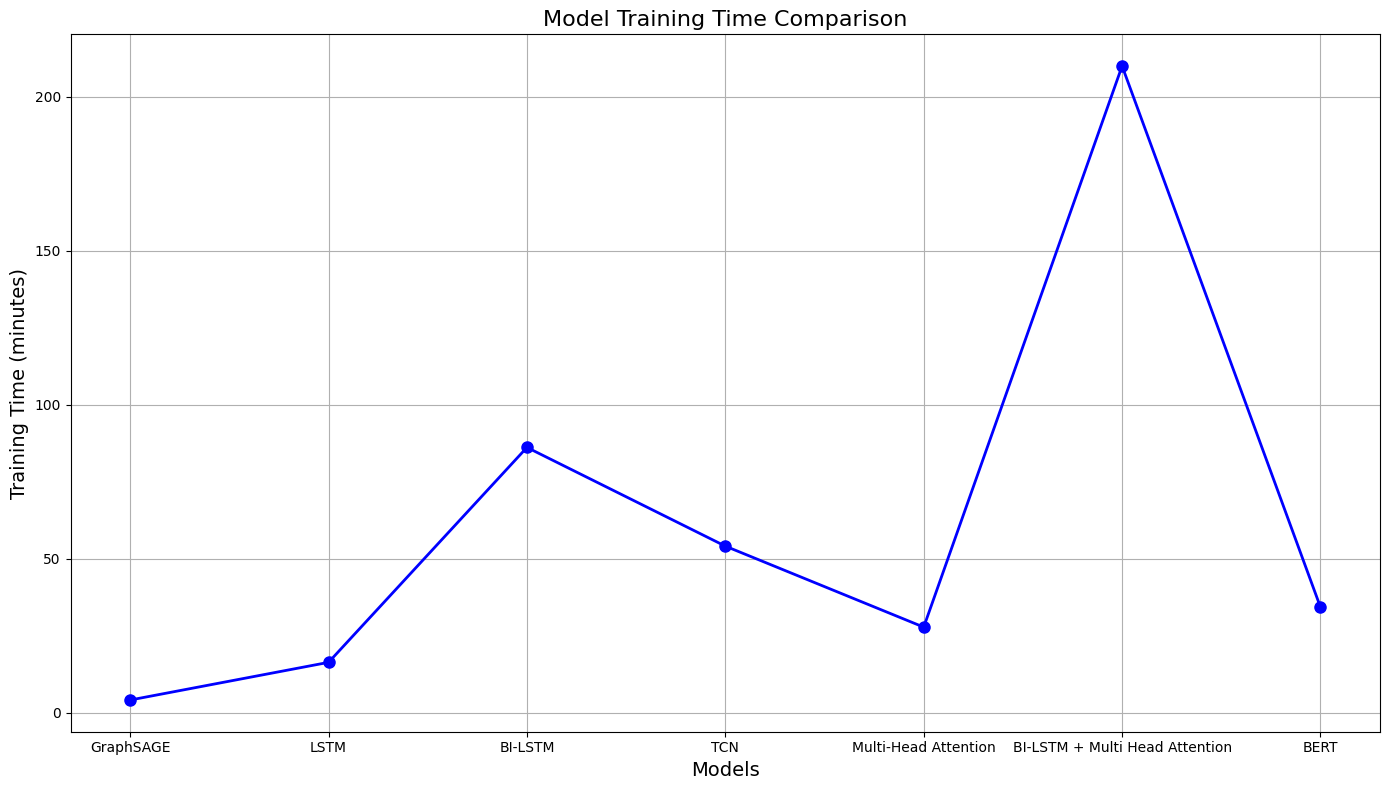}
    \caption{Training Time Comparison of Models}
    \label{fig:training_time}
    \vspace*{-0.20in}
\end{figure}

Figure \ref{fig:training_time} demonstrates the comparison of training times among different deep learning models.
GraphSAGE has the shortest training time of 4 minutes and 6 seconds. LSTM has the training times approximately 16 minutes and 2 seconds. On the other hand, BI-LSTM, which considers sequences bidirectionally, significantly increases the training time to 86 minutes 42 seconds. This is anticipated since there is additional computational expense of processing input sequences in both forward and backward directions.

TCN, as with the convolutional network, achieved a training moment time involving nearly 54  minutes, advantages from parallelism while keeping sequence-level dependencies. Multi-Head Attention receives a relatively manageable training time of about 27 minutes 48 seconds, attributed to its parallelized computations, and it is a well-suited option for various sequence tasks. On the other hand, the BERT model requires 34 minutes and 14 seconds to complete model training. The efficiency results from BERT's self-attention mechanism because such architecture performs parallelized computation across entire sequences.

Furthermore, BI-LSTM with Multi-Head Attention has a maximum training time of about 210 minutes. This is because of the extra complexity of combining bidirectional sequence learning with multiple auxiliary attention heads, resulting in a large computational cost. This study exposes the key trade-off between model complexity and training time, pointing out that computational resources should be taken into account when choosing the models to be used in real-time malware detection. The model training and experiments conducted  on the Apple M2 Pro with 32 GB of memory, utilizing the GPU.

\begin{figure*}[htb]
    \centering

    \begin{subfigure}[b]{0.22\linewidth}
        \centering
        
         \includegraphics[width=\linewidth]{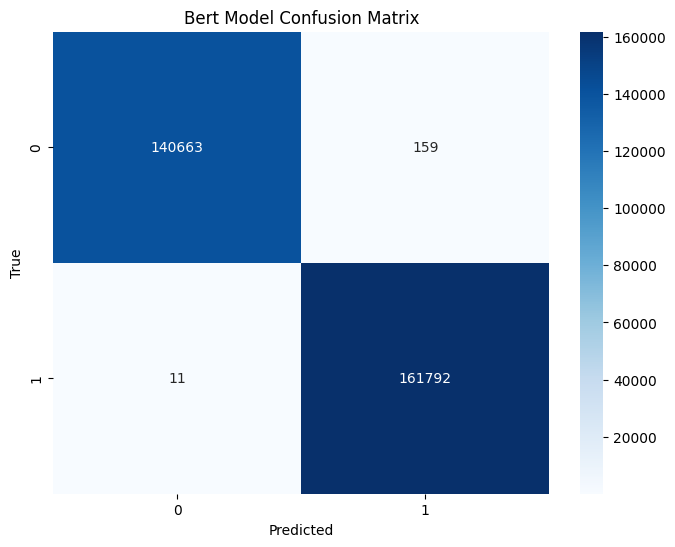}
        \caption{\footnotesize{BERT Model}}
        \label{fig:bert_conf}
    \end{subfigure}
    \hfill
    \begin{subfigure}[b]{0.22\linewidth}
        \centering
        \includegraphics[width=\linewidth]{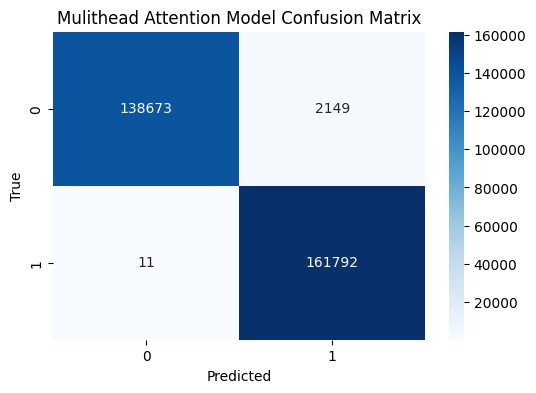}
        \caption{\footnotesize{Multi Head Attention Model}}
        \label{fig:multihead_conf}
    \end{subfigure}
    \hfill
    \begin{subfigure}[b]{0.22\linewidth}
        \centering
        \includegraphics[width=\linewidth]{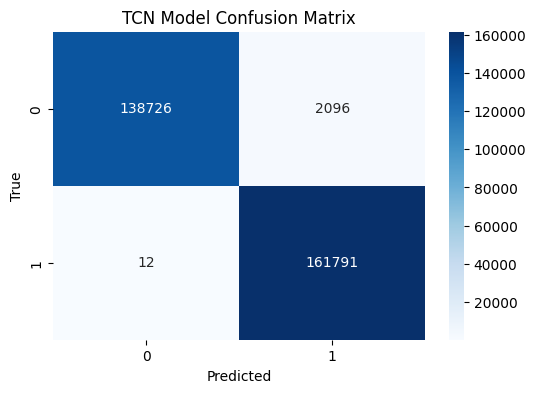}
        \caption{\footnotesize{TCN Model}}
        \label{fig:tcn_conf}
    \end{subfigure}
    \hfill
    \begin{subfigure}[b]{0.22\linewidth}
        \centering
        \includegraphics[width=\linewidth]{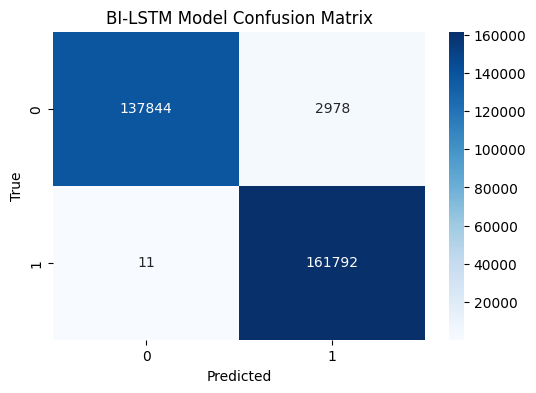}
        \caption{\footnotesize{BI-LSTM Model}}
        \label{fig:bilstm_conf}
    \end{subfigure}

    \vspace{0.6em}

    % Second row: 3 figures side-by-side
    \begin{subfigure}[b]{0.22\linewidth}
        \centering
        \includegraphics[width=\linewidth]{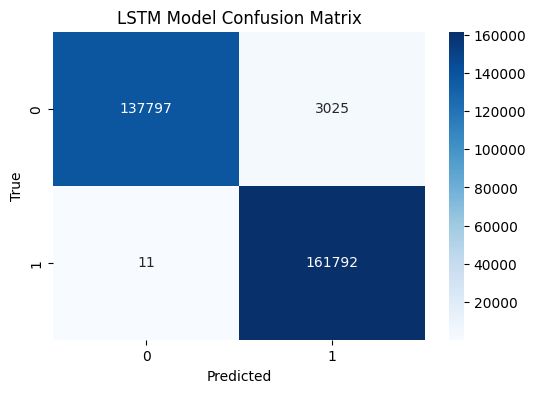}
        \caption{\footnotesize{LSTM Model}}
        \label{fig:lstm_conf}
    \end{subfigure}
    % \hfill
    \begin{subfigure}[b]{0.22\linewidth}
        \centering
       \includegraphics[width=\linewidth]{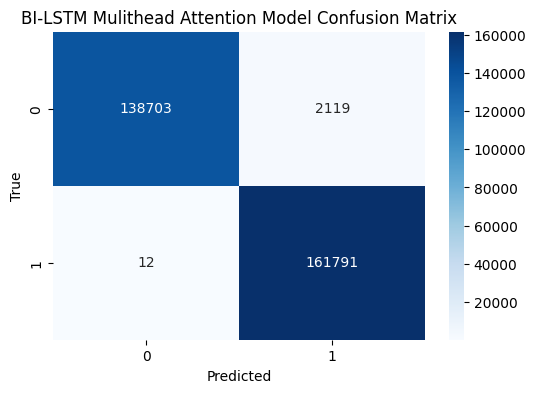}
        \caption{\footnotesize{BI-LSTM Multi-Head Attention Model}}
        \label{fig:bi_multi_conf}

    \end{subfigure}
    % \hfill
    \begin{subfigure}[b]{0.22\linewidth}
        \centering
        \includegraphics[width=\linewidth]{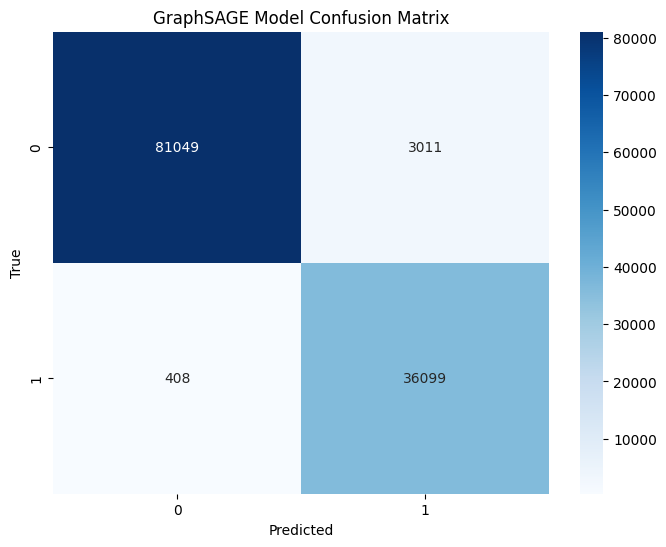}
        \caption{\footnotesize{GraphSAGE Model}}
        \label{fig:graphsage_conf}
    \end{subfigure}

    \caption{\footnotesize{Confusion Matrices of the evaluated models.}}
    \label{fig:confusion_fig}
    \vspace*{-0.20in}
\end{figure*}

\subsection{Confusion Matrix}

Figure \ref{fig:confusion_fig} presents the confusion matrix of the deep learning models from the experiment. The confusion matrix of LSTM model \ref{fig:lstm_conf} shows exceptional classification results that contain 137,797 true positives, 161,792 true negatives, 3,025 false positives together with 11 false negatives.  The BI-LSTM \ref{fig:bilstm_conf} achieved equivalent performance with results containing 137,844 true positives, 161,792 true negatives, 2,978 false positives and 11 false negatives. The implementation of past and future time steps contexts in the bidirectional approach enables the model to reduce false positives while keeping similar accuracy rates as compared to LSTM models.

The TCN \ref{fig:tcn_conf} outperforms both LSTM and BI-LSTM in terms of accuracy, with $138,726$ true positives, 161,791 true negatives, 2,096 false positives, and 12 false negatives. The TCN shows outstanding ability to accurately model time sequences through its convolutional structure which results in fewer false detection incidents thus establishing it as an optimal methodology for malware detection.

Furthermore, the Multi-Head Attention model \ref{fig:multihead_conf} detected $138,673$ true positives, $161,792$ true negatives and $2,149$ false positives and 11 false negatives. The attention mechanism of this model displays interpretability because it identifies important features in the input data although it produces slightly more false positives than the TCN model.

When BI-LSTM integrates Multi-Head Attention \ref{fig:bi_multi_conf} it succeeds in classifying $138,703$ correct samples while achieving 161,791 accurate negatives and marking 2,119 instances wrong. The model identifies 12 incorrect cases as well. The integrated architecture uses both network components to extract bidirectional patterns from data yet considers the most important features through attention-based processing. 
% The model's neutral results indicate it can serve as an optimal method to boost both accuracy and robustness in classification tasks.

BERT model \ref{fig:bert_conf} displays exceptional classification which detected $161,792$ true positives and $140,663$ true negatives together with 159 false positives and 11 false negatives. The robustness of BERT in malware detection is demonstrated by the noticeably decreased false positive rate when compared to other models. 

% BERT benefits from its self-attention capability and deep contextualized representation modeling that lets it identify malicious traffic patterns better than normal ones. BERT demonstrates great effectiveness as a model for minimizing false alarms in practical malware detection systems due to its low failure rates.

Moreover, the GraphSAGE model \ref{fig:graphsage_conf} attained low accuracy on classification, identifying $81,049$ samples to be class 0 (true negatives) and $36,099$ samples to be class 1 (true positives). Although $3,011$ samples classified as false positives, 408 samples mistakenly classified as false negatives. The model failed to detect the highest number in false positives and false negatives compared to other models.

\section{Conclusion}
\label{sec:conclusion}

This paper compared the performance of a graph-based model, GraphSAGE and deep learning models including BERT,  TCN, Multi Head Attention, BI-LSTM with Multihead Attention, BI-LSTM, and LSTM in identifying malicious network traffic. Experimental results show that while all models performed comparably, BERT outperforms all other models, and achieves the highest accuracy (99.94\%) and AUC-ROC (99.99\%). The TCN, Multihead Attention and BI-LSTM with Multi-Head Attention models also competitively highlight the value of attention mechanisms in dealing with temporal dependencies in network traffic. Although BI-LSTM and LSTM resulted in less than optimal performance, they still performed very well and showed that recurrent architectures is a good fit for sequential malware detection tasks. The GraphSAGE model has the lowest performance across all metrics compared to other models. In terms of training time, the GraphSAGE model is trained within  4 minutes and 6 seconds. BI-LSTM with Multi-Head Attention model takes the longest training time of 210 minutes. The experiment results revealed that model complexity creates an important competing relationship with computational performance during the training process.

% The convolutional design of TCN produced a middle-time training process of about 54 minutes which provided both accuracy retention and parallel computing advantages. The training process of BI-LSTM with Multi-Head Attention proved excessively time-consuming which makes it undesirable for real-time applications.

The confusion matrix evaluation confirmed BERT's high performance levels because it generated the lowest number of false positive detection results thus improving its trustworthiness for malware detection. The Multi-Head Attention along with BI-LSTM approaches yielded suitable results yet they led to elevated rates of false positives. The GraphSAGE model delivered the lowest accuracy and shortest training time, even though it produced the most false positive results.

The study establishes that BERT and TCN emerge as the superior selection for network traffic malware detection because it combine excellent accuracy with fast training performance and effective tracking of temporal relationships. Additional attention mechanisms enhance other models with readable interpretations and selection features yet require increased computational complexity. In the future, we would like to consider hybrid models that combine the best attributes of different architectures, like coupling the temporal dependency handling of transformer models like BERT with the computational efficiency of graph-based models like GraphSAGE. 

% Furthermore, making models to be time-efficient for real-time detection at the same time keeping robust performance remains a critical challenge. Model optimization work should focus on real-time applicability improvements without compromising detection performance.

\section*{Acknowledgment} 
This work is partially supported by a grant from the National Science Foundation (Award No. 2319802).

\bibliography{refs}{}
\bibliographystyle{plain}

\end{document}